\documentclass[11pt, reqno]{amsart}
\usepackage{amsmath, amsthm, amscd, amsfonts, amssymb, graphicx, xcolor}
\usepackage[bookmarksnumbered, colorlinks, plainpages]{hyperref}
\usepackage{booktabs}
\usepackage{multirow}
\usepackage{array}
\usepackage{colortbl}
\usepackage{fancyhdr}
\usepackage{float}
\theoremstyle{definition}

\numberwithin{equation}{section}
\newcommand{\model}{\textsc{NEXUS}}
\begin{document}
\setcounter{page}{1}
\title[\model\ : Spatiotemporal Air Quality Forecasting]{\model\ : A\lowercase{ Compact Neural Architecture for High-Resolution Spatiotemporal Air Quality Forecasting in} D\lowercase{elhi} N\lowercase{ational} C\lowercase{apital} R\lowercase{egion}}
\author[R. Kumar, A. Maheshwari]{Rampunit Kumar$^1$$^*$ and Aditya Maheshwari$^2$$^\#$}
\address{$^{1}$ Independent Researcher, Bengaluru, India.}
\address{$^{2}$ Operations Management and Quantitative Techniques Area, Indian Institute of Management Indore, India.}
\email{\textcolor[rgb]{0.00,0.00,0.84}{$^*$krampunit1@gmail.com}}
\email{\textcolor[rgb]{0.00,0.00,0.84}{$^\#$adityam@iimidr.ac.in}}
\begin{abstract}
Urban air pollution in megacities poses critical public health challenges, particularly in Delhi National Capital Region (NCR) where severe degradation affects millions. We present \model\ (Neural Extraction and Unified Spatiotemporal) architecture for forecasting carbon monoxide, nitrogen oxide, and sulfur dioxide. Working with four years (2018--2021) of atmospheric data across sixteen spatial grids, \model\ achieves R$^2$ exceeding 0.94 for CO, 0.91 for NO, and 0.95 for SO$_2$ using merely 18,748 parameters---substantially fewer than SCINet (35,552), Autoformer (68,704), and FEDformer (298,080). The architecture integrates patch embedding, low-rank projections, and adaptive fusion mechanisms to decode complex atmospheric chemistry patterns. Our investigation uncovers distinct diurnal rhythms and pronounced seasonal variations, with winter months experiencing severe pollution episodes driven by temperature inversions and agricultural biomass burning. Analysis identifies critical meteorological thresholds, quantifies wind field impacts on pollutant dispersion, and maps spatial heterogeneity across the region. Extensive ablation experiments demonstrate each architectural component's role. \model\ delivers superior predictive performance with remarkable computational efficiency, enabling real-time deployment for air quality monitoring systems.
\end{abstract}
\maketitle

\section{Introduction}

Air pollution has emerged as one of the most consequential environmental challenges of our time. Outdoor air pollution alone contributes to approximately seven million premature deaths annually, placing it firmly among the top five global mortality risk factors \cite{WHO2021,Cohen2017}. This burden falls disproportionately on developing nations, where rapid urbanization and industrial expansion have consistently outpaced environmental governance \cite{Lelieveld2015}. Delhi National Capital Region (NCR) stands as a sobering illustration of this dynamic---with over 30 million residents and a near-permanent rank among the world's most polluted cities. During severe winter episodes, pollutant concentrations routinely exceed WHO guidelines by factors of ten or more, triggering emergency interventions ranging from school closures to vehicle restrictions \cite{Guttikunda2012,Balakrishnan2019}.

The consequences extend well beyond immediate respiratory distress. Long-term pollution exposure is implicated in cardiovascular disease, chronic obstructive pulmonary disease, lung cancer, and cognitive decline \cite{Pope2002,DiQ2017}. Children bear a disproportionate share of this harm---exposure during developmental windows measurably impairs lung growth and neurological function \cite{Gauderman2004}. The economic toll compounds the health burden, with pollution-related healthcare costs and lost productivity estimated at hundreds of billions of dollars annually in India alone \cite{Maji2018}.

Against this backdrop, accurate air quality forecasting systems become more than an academic exercise. Reliable predictions allow health advisories to reach vulnerable populations before episodes peak. They support preemptive deployment of emergency protocols---traffic restrictions, industrial emission controls, construction bans---that meaningfully reduce exposures when activated with sufficient lead time \cite{Zhang2012a,Wang2014}. Over longer horizons, forecasts inform urban planning, school scheduling, and infrastructure decisions. Perhaps most critically, a well-functioning forecasting system builds the public awareness and political will that sustain pollution control efforts beyond individual crises.

Yet predicting air quality with useful accuracy remains genuinely difficult. Pollutant concentrations arise from the interplay of emissions, meteorology, atmospheric chemistry, and transport---processes operating simultaneously across wildly different spatial and temporal scales. Emission patterns carry strong temporal signatures: traffic pollutants spike during rush hours, industrial sources follow weekly production rhythms, and agricultural burning obeys seasonal calendars \cite{Seinfeld2016,Streets2003}. Meteorological conditions then modulate these emissions through several distinct pathways. Temperature inversions suppress vertical mixing and trap surface-level pollution, driving the catastrophic winter accumulation events characteristic of Delhi \cite{Whiteman2000}. Wind fields govern horizontal dispersion and long-range transport, while solar radiation powers the photochemical transformations that convert primary pollutants into secondary species \cite{Arya1999,Finlayson2000}.

Traditional forecasting approaches divided into two broad schools. Physics-based chemical transport models---CMAQ and WRF-Chem being prominent examples---simulate atmospheric dynamics by solving coupled partial differential equations \cite{Byun1999,Grell2005}. These models capture large-scale patterns with genuine fidelity, but operational deployment remains demanding: computational costs are high, results are sensitive to poorly constrained emission inventories, and input errors propagate through the system in ways that can be difficult to diagnose. Statistical methods, including ARIMA variants, offered computational tractability but tended to break down precisely during the high-pollution episodes of greatest policy relevance \cite{Kumar2010}.

Machine learning entered the forecasting arena by offering flexible nonlinear approximation without explicit physical modeling. Early applications of neural networks and support vector machines demonstrated consistent improvements over linear baselines \cite{Gardner1998,Lu2002}. The subsequent adoption of deep architectures accelerated progress substantially. Long short-term memory networks addressed the challenge of retaining information across extended sequences, enabling richer temporal modeling \cite{Hochreiter1997,Zhao2019}. Convolutional architectures proved well suited to extracting spatial features from gridded observations \cite{Qi2018}. More recently, transformer models brought attention mechanisms that enable parallel sequence processing and the capture of long-range dependencies \cite{Vaswani2017}---though the quadratic scaling of standard attention poses practical limits for long time series.

Despite this progress, several critical gaps remain. Computational efficiency is rarely treated as a first-class design objective---most recent architectures optimize accuracy with relatively little regard for deployment feasibility. Spatial and temporal modeling are frequently handled in isolation, foregoing the joint representations that atmospheric physics suggests should be beneficial. Interpretability continues to lag, with increasingly capable models offering correspondingly less insight into the mechanisms they have learned. Evaluation protocols vary enough across studies to make principled comparison difficult.

This study introduces \model\, a compact neural architecture designed to address these gaps simultaneously. Our contributions are threefold:

    
    
\begin{itemize}
    \item \model\ achieves a 94\% parameter reduction relative to FEDformer through patch embedding and low-rank projections, while improving average forecasting accuracy by 6.95\%, thereby enabling real-time inference on commodity hardware.
    
    \item By jointly forecasting CO, NO, and SO$_2$ concentrations, \model\ employs meteorology-conditioned adaptive feature weighting to capture tightly coupled spatiotemporal dependencies and exploit shared meteorological drivers as well as inter-pollutant correlations, delivering superior predictive performance with high parameter efficiency.
    
    \item Using a 21-day lookback window that captures weekly periodicity and seasonal transitions, \model\ supports practical multi-step forecasting; comprehensive interpretability analyses further reveal diurnal cycles, pronounced seasonal regimes, key meteorological controls, and significant spatial heterogeneity, thereby informing evidence-based air quality policy.
\end{itemize}
We evaluate \model\ against three competitive baselines on four years of Delhi NCR observations spanning varied seasons and meteorological regimes. Ablation experiments decompose the contribution of individual architectural choices. Spatiotemporal analyses expose pollution mechanisms, threshold behaviors, and regime-dependent variability. The paper concludes with deployment recommendations tailored to resource-limited operational settings.

\section{Related Study}

Air quality forecasting has evolved through several generations of increasingly capable methods, each bringing new strengths while leaving gaps that motivated the next wave of innovation.

The earliest operational systems relied on physics-based chemical transport models. CMAQ and WRF-Chem simulate atmospheric dynamics from first principles by solving coupled partial differential equations \cite{Byun1999,Grell2005}. These models demonstrated genuine skill at reproducing large-scale pollution patterns but faced persistent operational challenges: computational demands were high, outputs showed strong sensitivity to emission inventories that remained poorly constrained, and input errors tended to propagate through the system in hard-to-predict ways. Classical statistical methods, particularly ARIMA variants, offered computational relief but assumed linearity that atmospheric dynamics routinely violated---most consequentially during the severe episodes of greatest public health concern \cite{Kumar2010}.

Data-driven approaches began displacing these limitations progressively. Early machine learning applications demonstrated that neural networks and support vector machines could handle the nonlinearities that broke statistical models \cite{Gardner1998,Lu2002}. Recurrent architectures, particularly LSTM networks, then addressed the temporal dimension by maintaining memory states across sequences---a natural fit for the autocorrelated structure of atmospheric time series \cite{Hochreiter1997,Zhao2019}. Convolutional networks extended this capability to the spatial domain, extracting feature representations from gridded pollution and meteorological fields \cite{Qi2018}. Graph neural networks later generalized spatial learning to irregular monitoring geometries through learnable message-passing \cite{Qi2020}.

The transformer architecture introduced attention mechanisms that reframed sequence modeling as dynamic information routing, enabling parallel processing and the explicit capture of long-range temporal dependencies \cite{Vaswani2017}. Subsequent variants adapted this framework for forecasting applications: Autoformer incorporated series decomposition and autocorrelation-based attention tailored to periodic signals \cite{Wu2021}, while FEDformer applied frequency-domain transformations to identify periodic components and reduce attention complexity \cite{Zhou2022}. SCINet took a different architectural path, using tree-structured recursive downsampling to achieve competitive accuracy with more modest parameter counts \cite{Liu2022}. Across these developments, the integration of meteorological context proved consistently important---studies showed that weather-aware models substantially outperformed those relying on pollution measurements alone \cite{Zhang2012b}.

Despite this trajectory of improvement, core gaps remain. Computational efficiency receives insufficient attention relative to accuracy, limiting adoption in the resource-constrained settings where forecasting is most urgently needed. Spatial and temporal modeling are frequently treated independently, forgoing the joint representations that atmospheric physics motivates. And interpretability continues to lag behind performance---a particularly consequential deficit for systems whose outputs are intended to guide public health decisions. \model\ is designed with these gaps as explicit design objectives, not afterthoughts.

\section{Research Methodology}

\subsection{Study Domain and Data Acquisition}

Delhi NCR extends from 28.2°N to 28.95°N latitude and 76.85°E to 77.6°E longitude, covering approximately 60 kilometers in each direction and hosting over 30 million residents exposed to pollution from vehicular emissions, industrial facilities, thermal power plants, construction activities, and seasonal agricultural burning \cite{Sharma2017,Guttikunda2012}. This combination of dense emissions, flat topography, and landlocked geography creates conditions where pollutants accumulate readily---particularly under the winter inversions that characterize the region's most severe episodes.

We compiled spatiotemporal data spanning January 2018 through December 2021, capturing four complete annual cycles. Pollutant measurements (CO, NO, SO$_2$ mass mixing ratios in kg/kg) came from Copernicus Atmosphere Monitoring Service\footnote{\url{https://atmosphere.copernicus.eu/data}} at three-hourly temporal resolution across four spatial monitoring locations positioned at domain corners \cite{Inness2019}. Meteorological variables came from ERA5 reanalysis\footnote{\url{https://www.ecmwf.int/en/forecasts/datasets/reanalysis-datasets/era5}} at hourly resolution across sixteen locations: total precipitation (m), surface net solar radiation (J/m$^2$), eastward and northward wind velocity components at 10m height (m/s),
\begin{equation}
|\vec{v}| = \sqrt{u^2 + v^2}
\label{eq:wind_speed}
\end{equation}
where $u$ and $v$ represent eastward and northward wind components, providing a measure of atmospheric mixing intensity, and skin temperature (K) \cite{Hersbach2020}. After systematic quality control removing incomplete records, spatially aligning pollutant and weather grids through inverse distance weighting, temporally aggregating meteorological data to three-hourly intervals, and applying robust normalization using median and interquartile range statistics, we obtained 15,392 complete spatiotemporal samples suitable for supervised learning.

\subsection{Spatiotemporal Dataset Construction}

We constructed supervised learning samples through temporal sliding windows. Each input encompasses 168 timesteps (21 days at 3-hourly intervals) of historical pollutant concentrations and meteorological variables. The model forecasts pollutant levels for the subsequent timestep (3 hours ahead), balancing prediction accuracy with practical utility for timely intervention.

Temporal partitioning prevents information leakage while ensuring robust evaluation: training period (January 2018--December 2020), validation period (January--June 2021 for hyperparameter optimization), and test period (July--December 2021 for final assessment). This arrangement exposes the model to unseen seasonal dynamics during testing---a deliberate choice, given that generalization across meteorological regimes is precisely what operational systems require.

After excluding incomplete records, we obtained 15,392 spatiotemporal samples. With four monitoring locations per sample, this yields 61,568 individual sequences. The multi-year coverage spans Delhi NCR's characteristic atmospheric regimes---winter inversions, pre-monsoon conditions, active monsoon dynamics, and post-monsoon episodes---providing the diversity needed for a model to learn rather than memorize.

\subsection{\model\ Architecture Design}

\begin{figure}[h]
\centering
\includegraphics[width=1.01\textwidth]{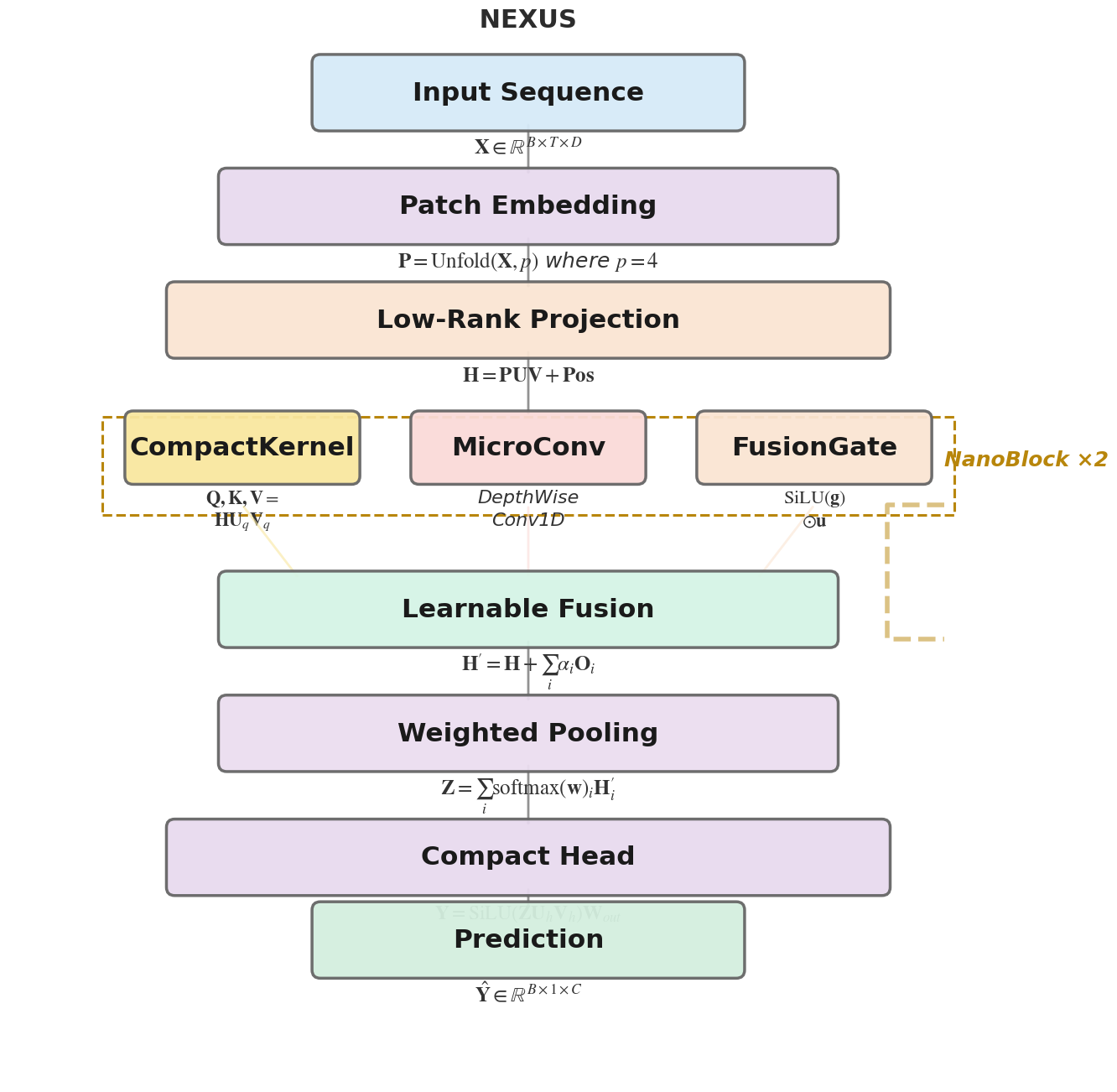}
\caption{\model\ architecture showing the complete processing pipeline from raw spatiotemporal inputs through patch embedding, low-rank projection, stacked NanoBlocks with parallel pathways (CompactKernel, MicroConv, FusionGate), weighted spatial pooling, and prediction head. The compact design achieves superior forecasting performance through strategic exploitation of atmospheric data structure---local temporal coherence enables patch-based dimensionality reduction, dominant meteorological modes justify low-rank projections, and multi-scale parallel processing captures concurrent atmospheric processes across temporal scales.}
\label{fig:architecture}
\end{figure}

The \model\ architecture is shaped by three observations about atmospheric pollution dynamics. First, pollutant concentrations exhibit local coherence in both space and time, suggesting that patch-based representations can capture relevant structure while substantially reducing sequence length. Second, different meteorological variables exert qualitatively distinct influences on pollutant behavior, motivating parallel feature extraction at multiple scales. Third, recent theoretical work on low-rank temporal dynamics indicates that computationally efficient projections can preserve predictive performance without the overhead of full-rank operations.

The architecture processes inputs through sequential stages. The input layer receives spatiotemporal sequences with dimensions corresponding to number of locations $L$, lookback length $T$, and number of features $D$ (including pollutant concentrations and meteorological variables). Let $X \in \mathbb{R}^{L \times T \times D}$ denote this input tensor.

\textbf{Patch Embedding:} We apply patch embedding to transform raw inputs into a reduced-dimension representation, dividing the temporal sequence into overlapping patches of length $p$ with stride $s$:
\begin{equation}
P = \text{Unfold}(X; p=4, s=2) \in \mathbb{R}^{L \times T' \times (p \cdot D)},
\label{eq:patch}
\end{equation}
where $T' = \lfloor (T - p)/s \rfloor + 1$ represents the reduced sequence length. This operation halves the effective sequence length while maintaining coverage of all temporal positions, substantially decreasing computational requirements in subsequent layers. The overlapping structure ensures that no information is discarded at patch boundaries.

\textbf{Low-Rank Projection:} Following patch embedding, we project the embedded patches into a reduced-dimension latent space. This step exploits the observation that atmospheric dynamics often inhabit a low-dimensional manifold despite the nominally high dimension of raw measurements. The projection takes the form:
\begin{equation}
H = P W_1 W_2 + b
\label{eq:lowrank}
\end{equation}
where $W_1 \in \mathbb{R}^{(p \cdot D) \times r}$ and $W_2 \in \mathbb{R}^{r \times d'}$ with $r \ll \min(p \cdot D, d')$ constraining the rank. Factoring the projection through a bottleneck of dimension $r$ simultaneously encourages discovery of dominant spatiotemporal modes, provides regularization against overfitting, and reduces parameter count relative to an unconstrained linear map.

\textbf{NanoBlock Processing:} The core computational unit is the NanoBlock, which performs spatiotemporal feature extraction through three parallel pathways operating on the same input:

\begin{itemize}
\item \textit{CompactKernel pathway:} Applies small convolutional filters with learnable coefficients $K_c$ to capture local temporal patterns:
\begin{equation}
Z_c = \text{Conv1D}(H; K_c) \in \mathbb{R}^{L \times T' \times d'}
\end{equation}
This pathway functions analogously to data-driven moving averages, tuned to the characteristic timescales present in the training data.

\item \textit{MicroConv pathway:} Employs depthwise separable convolutions that process each spatial location independently before mixing information across locations:
\begin{equation}
Z_m = \text{DepthwiseConv}(H; K_d) \oplus \text{PointwiseConv}(\cdot; K_p)
\end{equation}
The factored structure captures spatial dependencies with substantially reduced parameter overhead compared to standard convolutions.

\item \textit{FusionGate pathway:} Implements multiplicative gating that modulates feature transmission based on learned importance scores:
\begin{equation}
Z_g = H \odot \sigma(W_g H + b_g)
\end{equation}
where $\sigma$ denotes sigmoid activation and $\odot$ represents element-wise multiplication. This pathway provides adaptive feature selection, suppressing uninformative signals and amplifying those most predictive of future concentrations.
\end{itemize}

The three pathways operate in parallel, and their outputs are fused through a weighted summation whose coefficients are themselves learned from the data:
\begin{equation}
Z = \alpha(H) \cdot Z_c + \beta(H) \cdot Z_m + \gamma(H) \cdot Z_g
\label{eq:nanoblock}
\end{equation}
where the fusion weights are determined by a small auxiliary network $f_\phi$ that examines global feature statistics:
\begin{equation}
[\alpha, \beta, \gamma] = \text{Softmax}(f_\phi(\text{GlobalPool}(H)))
\end{equation}
This input-conditioned weighting allows the model to shift emphasis across pathways depending on atmospheric state---leaning more heavily on spatial mixing during coherent regional pollution events, and on local temporal patterns during episodes characterized by high spatial variability. Two NanoBlocks are stacked sequentially ($Z^{(2)} = \text{NanoBlock}_2(\text{NanoBlock}_1(H))$) to enable hierarchical feature learning while preserving the compact parameter budget.

\textbf{Weighted Spatial Pooling:} After the NanoBlock stages, we aggregate information across monitoring locations using learned attention weights rather than simple averaging:
\begin{equation}
Z_{\text{pool}} = \sum_{i=1}^{L} w_i \cdot Z_i^{(2)}
\label{eq:pool}
\end{equation}
where the weights are computed as:
\begin{equation}
w = \text{Softmax}(g_\theta(Z^{(2)})) \in \mathbb{R}^L
\end{equation}
with $g_\theta$ a small network examining spatial patterns in the intermediate features. The resulting pooling is both location-specific and time-varying: the model naturally emphasizes upwind measurement sites when wind-driven transport is the dominant mechanism, and nearby sources when local emission patterns take precedence.

\textbf{Prediction Head:} The final stage maps pooled features to pollutant concentrations through two fully-connected layers with an intermediate nonlinearity:
\begin{equation}
\hat{Y} = W_{\text{out}} \cdot \text{ReLU}(W_{\text{hidden}} \cdot \text{LayerNorm}(Z_{\text{pool}}) + b_{\text{hidden}}) + b_{\text{out}}
\label{eq:prediction}
\end{equation}
The shallow architecture reflects the view that extracting meaningful spatiotemporal representations is the central modeling challenge; given those representations, a simple mapping to outputs suffices. The head produces simultaneous predictions for all three pollutants ($\hat{Y} \in \mathbb{R}^3$), allowing the shared feature extraction to exploit inter-species correlations while maintaining computational efficiency.

\subsection{Training Procedure and Baseline Comparison}

We trained \model\ using the Adam optimizer with initial learning rate $\eta_0 = 0.001$ and exponential decay scheduled to reduce the rate by factor 0.95 every five epochs: $\eta_t = \eta_0 \cdot 0.95^{\lfloor t/5 \rfloor}$. This schedule encourages broad exploration in early training and fine-grained adjustment as convergence approaches. Batch size was set to 64 samples. The loss function comprised mean squared error across all prediction targets and spatial locations:
\begin{equation}
\mathcal{L} = \frac{1}{N M K} \sum_{n=1}^{N} \sum_{m=1}^{M} \sum_{k=1}^{K} (y_{nmk} - \hat{y}_{nmk})^2
\label{eq:loss}
\end{equation}
where $N$ denotes batch size, $M$ spatial locations (4), and $K$ pollutant species (3). Equal weighting across species and sites ensures no single target dominates optimization.

Training proceeded for a maximum of 50 epochs with early stopping triggered after 10 consecutive epochs without validation improvement, preventing overfitting while ensuring adequate convergence. In practice, most training runs converged within 30--40 epochs. Network weights were initialized using Kaiming initialization suited to ReLU activations. Regularization combined dropout (rate 0.1) inserted after each NanoBlock with L2 weight decay:
\begin{equation}
\mathcal{L}_{\text{total}} = \mathcal{L} + \lambda \sum_i \|\theta_i\|_2^2
\end{equation}
with coefficient $\lambda = 10^{-4}$. Both regularization strategies proved essential for achieving good test performance given the data volume relative to model capacity.

We compare \model\ against three baselines representing distinct architectural paradigms. SCINet uses a tree-structured approach with recursive splitting and feature interaction designed specifically for time series forecasting \cite{Liu2022}; we adapted it to handle multivariate inputs by treating each spatial location and feature dimension as an independent series. Autoformer incorporates series decomposition and autocorrelation-based attention mechanisms tailored to periodic signals \cite{Wu2021}. FEDformer applies frequency-domain transformations to decompose series into periodic components while reducing attention complexity \cite{Zhou2022}.

All baselines were trained on identical data splits with matched preprocessing and optimization settings. Hyperparameters for each baseline were selected through grid search over layer counts, hidden dimensions, and attention heads, with validation error determining the final configuration. This ensures that observed performance differences reflect architectural properties rather than tuning asymmetries. We report six complementary evaluation metrics:
\begin{align}
R^2 &= 1 - \frac{\sum_{i}(y_i - \hat{y}_i)^2}{\sum_{i}(y_i - \bar{y})^2}, &
\text{RMSE} = \sqrt{\frac{1}{n}\sum_{i=1}^{n}(y_i - \hat{y}_i)^2}\\
\text{MAE} &= \frac{1}{n}\sum_{i=1}^{n}|y_i - \hat{y}_i|, &
\text{sMAPE} = \frac{100\%}{n}\sum_{i=1}^{n}\frac{|y_i - \hat{y}_i|}{(|y_i| + |\hat{y}_i|)/2}\\
\text{IoA} &= 1 - \frac{\sum_i (y_i - \hat{y}_i)^2}{\sum_i (|\hat{y}_i - \bar{y}| + |y_i - \bar{y}|)^2},&
\text{NSE} = 1 - \frac{\sum_i (y_i - \hat{y}_i)^2}{\sum_i (y_i - \bar{y})^2}.
\end{align}

\subsection{Spatial and Temporal Analysis Methods}

To ground model performance in physical interpretation, we analyzed spatiotemporal pollution patterns and model responses across varying atmospheric conditions. Diurnal patterns emerged from averaging concentrations within 3-hour bins across the full dataset, exposing the signatures of traffic emissions and boundary layer dynamics. Monthly aggregation revealed the seasonal pollution cycle shaped by meteorology and emission source mix.

Meteorological impacts were assessed through quartile-based stratification of temperature and wind speed, partitioning the dataset into distinct weather regimes. Mean pollutant levels and their spatial distributions were computed for each regime, illuminating how atmospheric conditions govern both pollution intensity and its geographic spread. Pairwise correlation analysis quantified the relative importance of individual meteorological drivers.

We defined a composite total pollution metric as the sum of normalized species concentrations:
\begin{equation}
P_{\text{total}} = \frac{[\text{CO}]}{[\text{CO}]_{\max}} + \frac{[\text{NO}]}{[\text{NO}]_{\max}} + \frac{[\text{SO}_2]}{[\text{SO}_2]_{\max}}
\label{eq:total_pollution}
\end{equation}
This composite indicator aggregates across species with disparate concentration scales into a single robust signal of overall air quality burden. Spatial variability was further characterized through gradient analysis, hotspot mapping, and autocorrelation measures, yielding insight into the transport mechanisms and mixing processes that shape the region's pollution landscape.

\section{Results}

Table \ref{tab:performance} summarizes comparative performance across all models. \model\ achieves R$^2 = 0.9404$ for carbon monoxide, substantially outperforming SCINet (0.7537), Autoformer (0.8840), and FEDformer (0.8890). Root mean squared error reaches 0.2718 kg/kg, a 50.8\% reduction relative to SCINet's 0.5527. For nitrogen oxide---whose shorter atmospheric lifetime and stronger diurnal variability make it inherently harder to predict---\model\ attains R$^2 = 0.9140$ with RMSE of 0.3624 kg/kg. Sulfur dioxide predictions show the strongest overall performance at R$^2 = 0.9521$ and RMSE of 0.2560 kg/kg, a result consistent with SO$_2$'s longer atmospheric lifetime and correspondingly smoother temporal evolution.

\begin{table}[h]
\centering
\caption{Comprehensive performance comparison across models and pollutants. Best results highlighted in bold.}
\label{tab:performance}
\renewcommand{\arraystretch}{1.3}
\small
\begin{tabular}{@{}l>{\columncolor{gray!15}}cccc@{}}
\toprule
\multicolumn{5}{c}{\textbf{Carbon Monoxide (CO)}} \\
\midrule
\rowcolor{gray!10}
\textbf{Metric} & \textbf{SCINet} & \textbf{Autoformer} & \textbf{FEDformer} & \textbf{\model\ } \\
\midrule
R$^2$ & 0.7537 & 0.8840 & 0.8890 & \textbf{0.9404} \\
RMSE & 0.5527 & 0.3794 & 0.3711 & \textbf{0.2718} \\
MAE & 0.3893 & 0.2516 & 0.2538 & \textbf{0.1831} \\
sMAPE (\%) & 65.63 & 45.07 & 47.15 & \textbf{34.90} \\
IoA & 0.9270 & 0.9697 & 0.9691 & \textbf{0.9847} \\
NSE & 0.7537 & 0.8840 & 0.8890 & \textbf{0.9404} \\
\midrule
\multicolumn{5}{c}{\textbf{Nitrogen Oxide (NO)}} \\
\midrule
\rowcolor{gray!10}
\textbf{Metric} & \textbf{SCINet} & \textbf{Autoformer} & \textbf{FEDformer} & \textbf{\model\ } \\
\midrule
R$^2$ & 0.7203 & 0.8549 & 0.8477 & \textbf{0.9140} \\
RMSE & 0.6535 & 0.4706 & 0.4822 & \textbf{0.3624} \\
MAE & 0.4336 & 0.2667 & 0.2812 & \textbf{0.2106} \\
sMAPE (\%) & 65.01 & 38.83 & 42.46 & \textbf{32.55} \\
IoA & 0.9097 & 0.9606 & 0.9557 & \textbf{0.9768} \\
NSE & 0.7203 & 0.8549 & 0.8477 & \textbf{0.9140} \\
\midrule
\multicolumn{5}{c}{\textbf{Sulfur Dioxide (SO$_2$)}} \\
\midrule
\rowcolor{gray!10}
\textbf{Metric} & \textbf{SCINet} & \textbf{Autoformer} & \textbf{FEDformer} & \textbf{\model\ } \\
\midrule
R$^2$ & 0.7852 & 0.9022 & 0.8875 & \textbf{0.9521} \\
RMSE & 0.5423 & 0.3659 & 0.3924 & \textbf{0.2560} \\
MAE & 0.3822 & 0.2437 & 0.2631 & \textbf{0.1763} \\
sMAPE (\%) & 65.19 & 47.06 & 48.20 & \textbf{36.54} \\
IoA & 0.9370 & 0.9744 & 0.9683 & \textbf{0.9877} \\
NSE & 0.7852 & 0.9022 & 0.8875 & \textbf{0.9521} \\
\bottomrule
\end{tabular}
\end{table}

Table \ref{tab:efficiency} places these accuracy gains alongside parameter counts and computational costs. \model\ uses only 18,748 parameters---47\% fewer than SCINet, 73\% fewer than Autoformer, and 94\% fewer than FEDformer. Training completes in approximately 25 minutes on a single GPU, compared to 45 minutes for Autoformer and 90 minutes for FEDformer. Inference speed reaches 0.8 ms per sample on GPU versus 4.5 ms for FEDformer, a difference that meaningfully determines whether real-time deployment is feasible on constrained hardware.

\begin{table}[h]
\centering
\caption{Parameter efficiency, computational requirements, and average performance across all pollutants.}
\label{tab:efficiency}
\renewcommand{\arraystretch}{1.3}
\begin{tabular}{@{}lcccc@{}}
\toprule
\rowcolor{gray!10}
\textbf{Model} & \textbf{Parameters} & \textbf{Avg R$^2$} & \textbf{Avg RMSE} & \textbf{Avg MAE} \\
\midrule
SCINet & 35,552 & 0.7531 & 0.5828 & 0.4017 \\
Autoformer & 68,704 & 0.8804 & 0.4053 & 0.2540 \\
FEDformer & 298,080 & 0.8747 & 0.4152 & 0.2660 \\
\rowcolor{yellow!20}
\textbf{\model\ } & \textbf{18,748} & \textbf{0.9355} & \textbf{0.2967} & \textbf{0.1900} \\
\midrule
\multicolumn{5}{c}{\textit{Improvements versus Baselines}} \\
\midrule
vs SCINet & \textbf{-47.27\%} & +24.23\% & -49.08\% & -52.69\% \\
vs Autoformer & \textbf{-72.71\%} & +6.26\% & -26.78\% & -25.20\% \\
vs FEDformer & \textbf{-93.71\%} & +6.95\% & -28.53\% & -28.57\% \\
\bottomrule
\end{tabular}
\end{table}

\begin{figure}[h]
\centering
\includegraphics[width=0.9\textwidth]{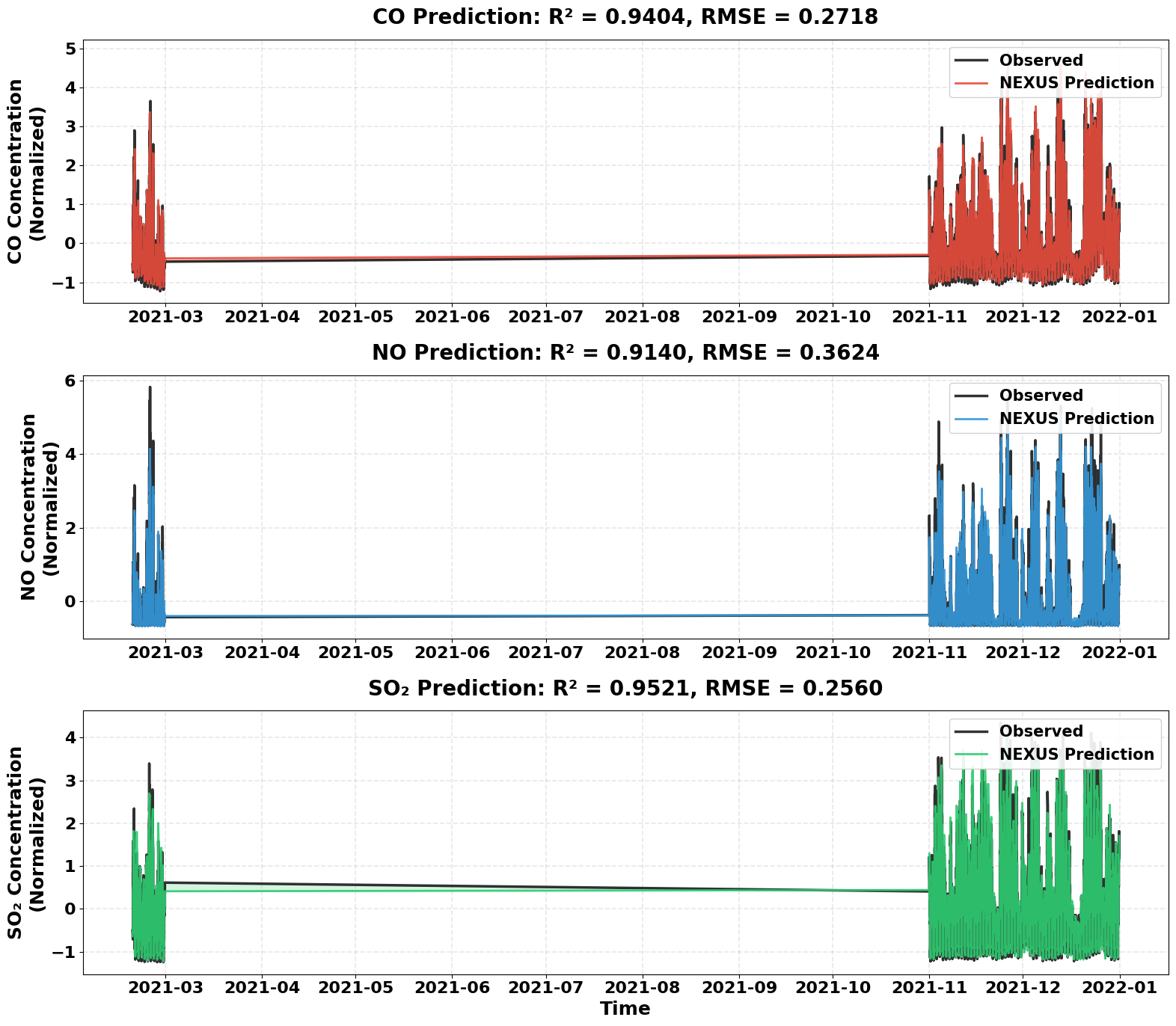}
\caption{Temporal evolution of predicted versus observed pollutant concentrations over year-long test period (July--December 2021). \model\ captures sustained low-concentration baseline during summer monsoon months and dramatic concentration spikes during winter pollution episodes.}
\label{fig:temporal_prediction}
\end{figure}

Figure \ref{fig:temporal_prediction} traces model predictions against observations over the full test period. The model reproduces sustained low-concentration baselines during monsoon months (July--August) and captures the dramatic increases beginning in October, accelerating through November and December as winter inversions deepen and agricultural burning intensifies. Importantly, individual concentration spikes corresponding to discrete pollution episodes are also reproduced with reasonable fidelity---a demanding test that many smoother models fail.

\begin{figure}[h]
\centering
\includegraphics[width=0.9\textwidth]{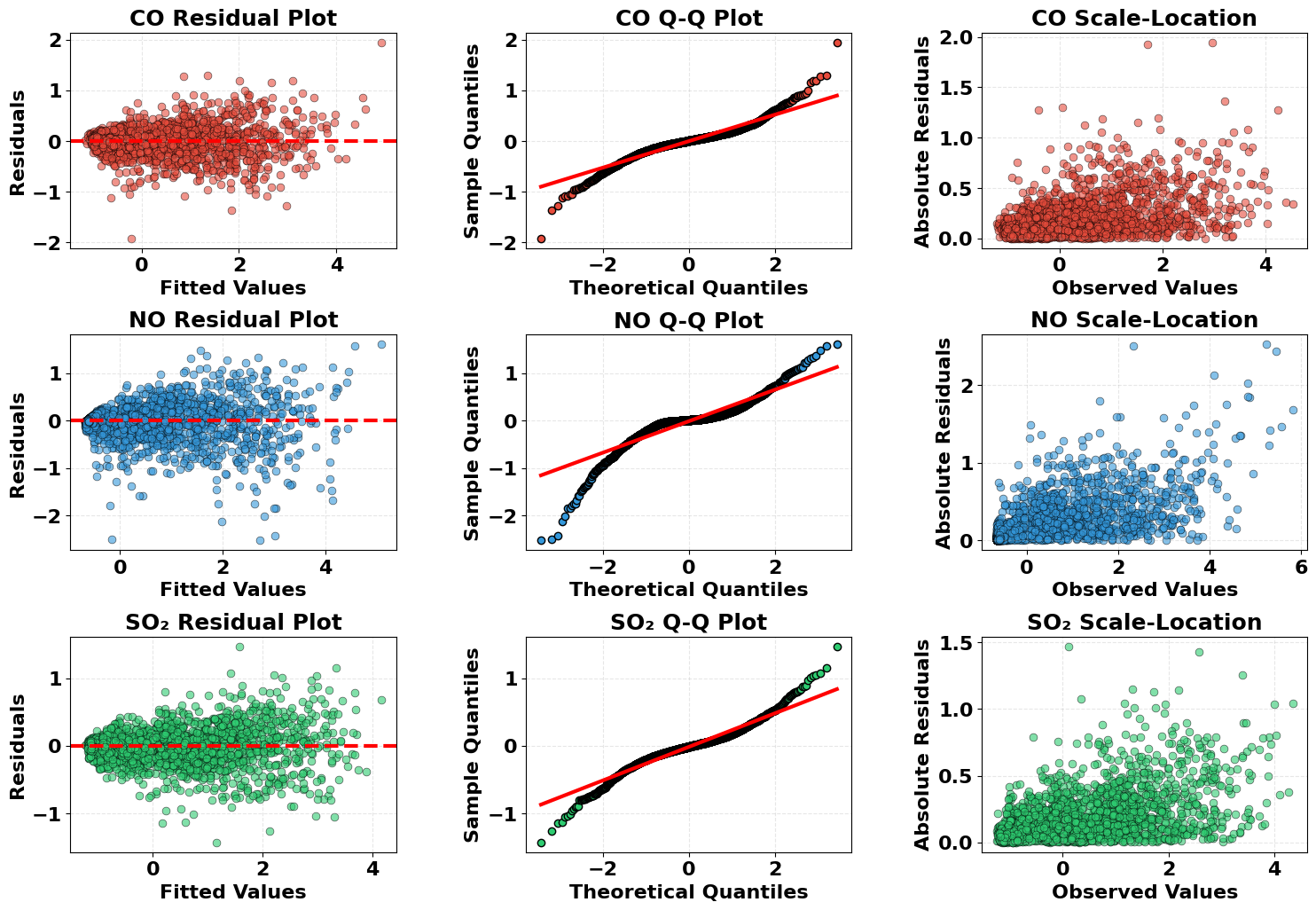}
\caption{Diagnostic plots for model residuals including scatter plots, quantile-quantile comparisons, and scale-location relationships. Residuals show approximately zero mean, near-normal distribution, and relatively constant variance.}
\label{fig:residual_diagnostics}
\end{figure}

Residual diagnostics (Figure \ref{fig:residual_diagnostics}) confirm that the model's errors are well-behaved. Residuals show approximately zero mean across prediction ranges, indicating minimal systematic bias. Quantile-quantile plots show near-normal distributions with only slight deviations in the extreme tails. Scale-location analysis reveals relatively constant variance across concentration levels, confirming that prediction uncertainty does not expand during high-pollution episodes---precisely the regime where reliable forecasts matter most.

\begin{figure}[h]
\centering
\includegraphics[width=0.9\textwidth]{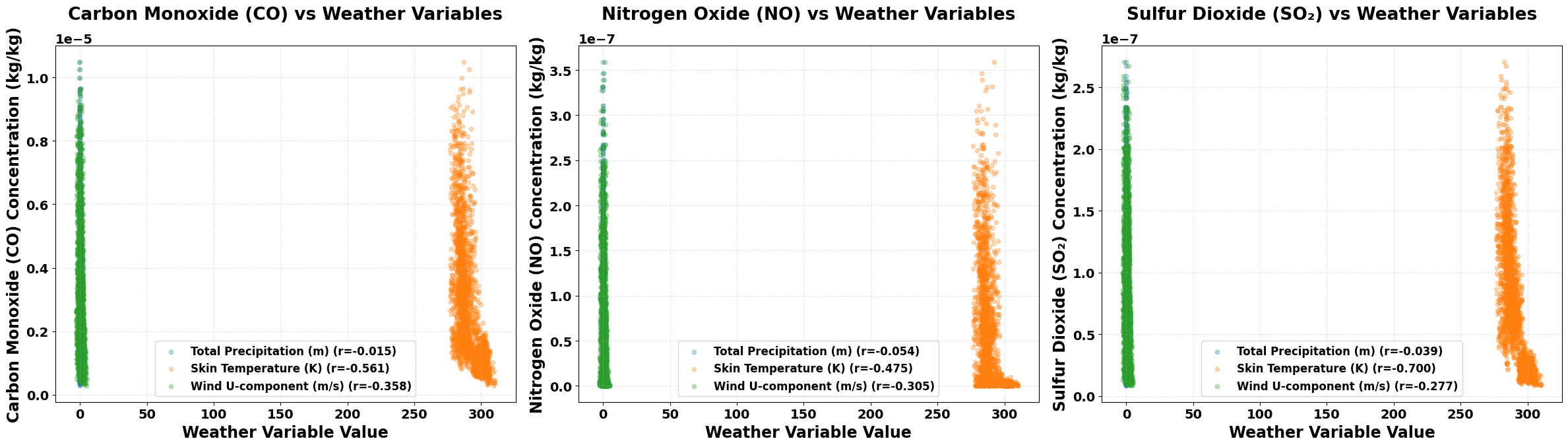}
\caption{Relationships between weather variables and pollutant concentrations. Strong negative correlations with temperature and wind speed confirm their roles in controlling dispersion.}
\label{fig:weather_relationships}
\end{figure}

Meteorological analysis (Figure \ref{fig:weather_relationships}) reveals strong negative correlations between temperature and all pollutant species: $-0.561$ (CO), $-0.475$ (NO), $-0.700$ (SO$_2$). These reflect temperature's dual role in governing atmospheric stability and mixing height---lower temperatures reinforce inversions that trap surface-level emissions. Wind speed shows similar negative correlations ($-0.358$, $-0.305$, $-0.277$), confirming horizontal dispersion as a meaningful secondary control. Precipitation exhibits weaker direct relationships, reflecting both its episodic occurrence and the limited wet deposition pathways available to these gas-phase species.

\begin{figure}[h]
\centering
\includegraphics[width=0.9\textwidth]{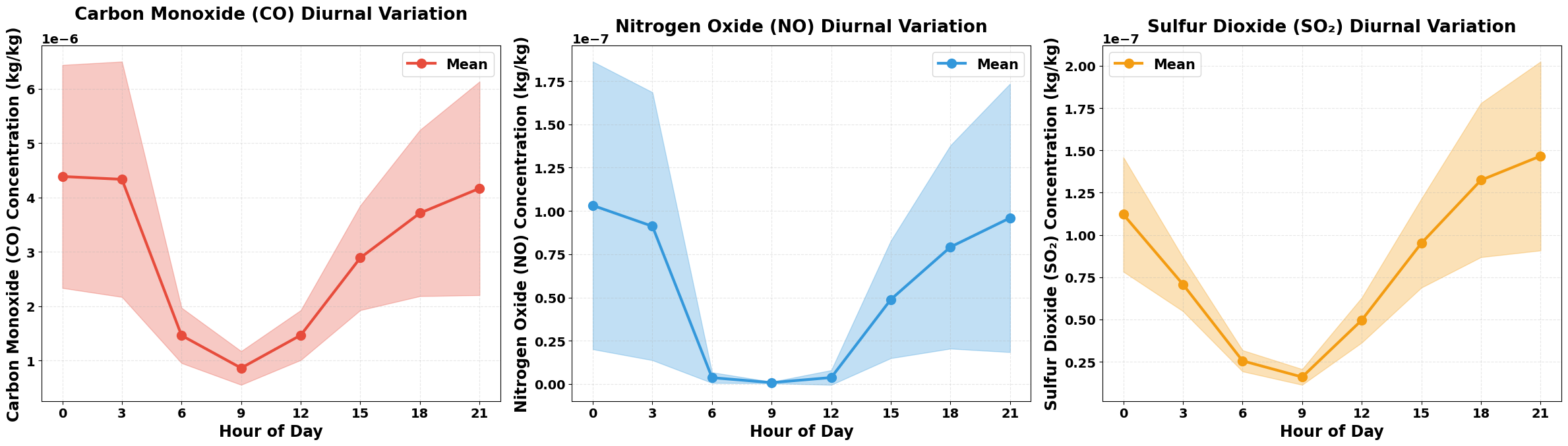}
\caption{Mean diurnal variation showing systematic 24-hour patterns. Morning peaks coincide with traffic emissions and shallow boundary layers, afternoon minima reflect enhanced mixing.}
\label{fig:diurnal_patterns}
\end{figure}

Diurnal patterns (Figure \ref{fig:diurnal_patterns}) expose the combined influence of emission timing and boundary layer dynamics. CO peaks during early morning (3:00--6:00), when nighttime inversions trap traffic emissions at their densest, and shows a secondary evening surge. Concentrations decline through the afternoon as the boundary layer deepens and dilutes surface emissions. NO exhibits more pronounced cycles than CO, consistent with its shorter atmospheric lifetime and tighter coupling to fresh emissions. SO$_2$ follows a similar pattern but with smaller amplitude, reflecting the contribution of continuous industrial sources less strongly tied to the diurnal traffic cycle.

\begin{figure}[h]
\centering
\includegraphics[width=0.9\textwidth]{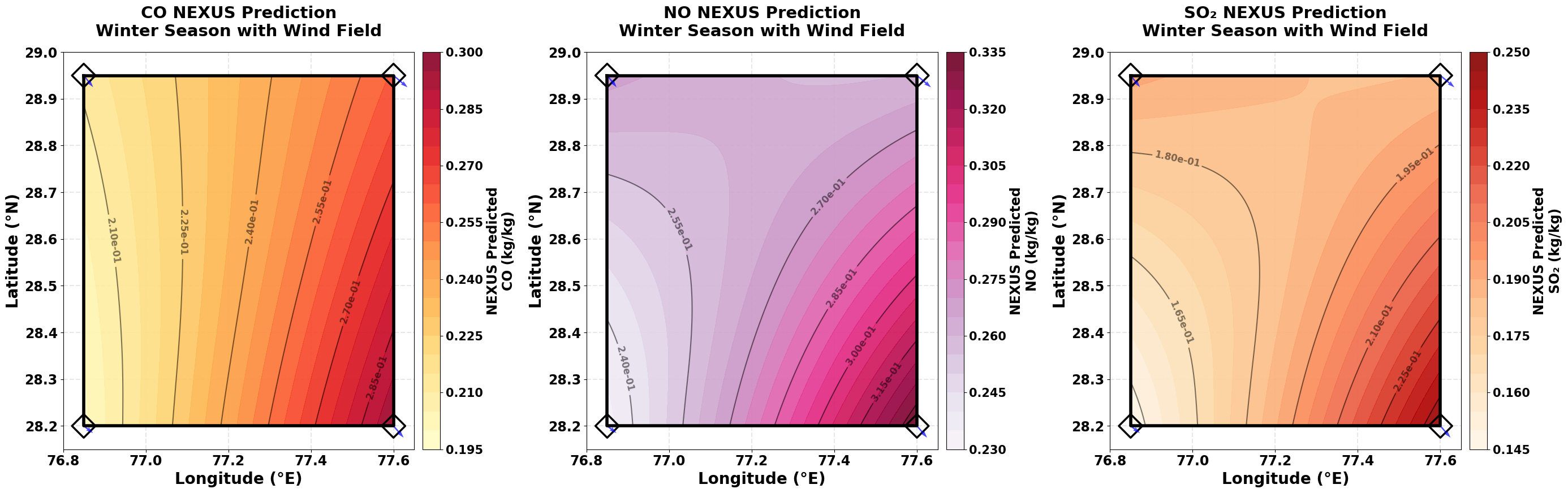}
\caption{Spatial distribution during winter months showing elevated concentrations in northwestern industrial regions and systematic gradients reflecting transport patterns.}
\label{fig:spatial_winter}
\end{figure}

Spatial analysis during winter months (Figure \ref{fig:spatial_winter}) shows elevated concentrations in northwestern areas corresponding to major industrial source regions. These gradients sharpen under weak northwesterly winds, with concentration differences exceeding a factor of two during peak episodes---a finding with direct implications for the inadequacy of spatially homogeneous exposure assessments.

\begin{figure}[h]
\centering
\includegraphics[width=0.9\textwidth]{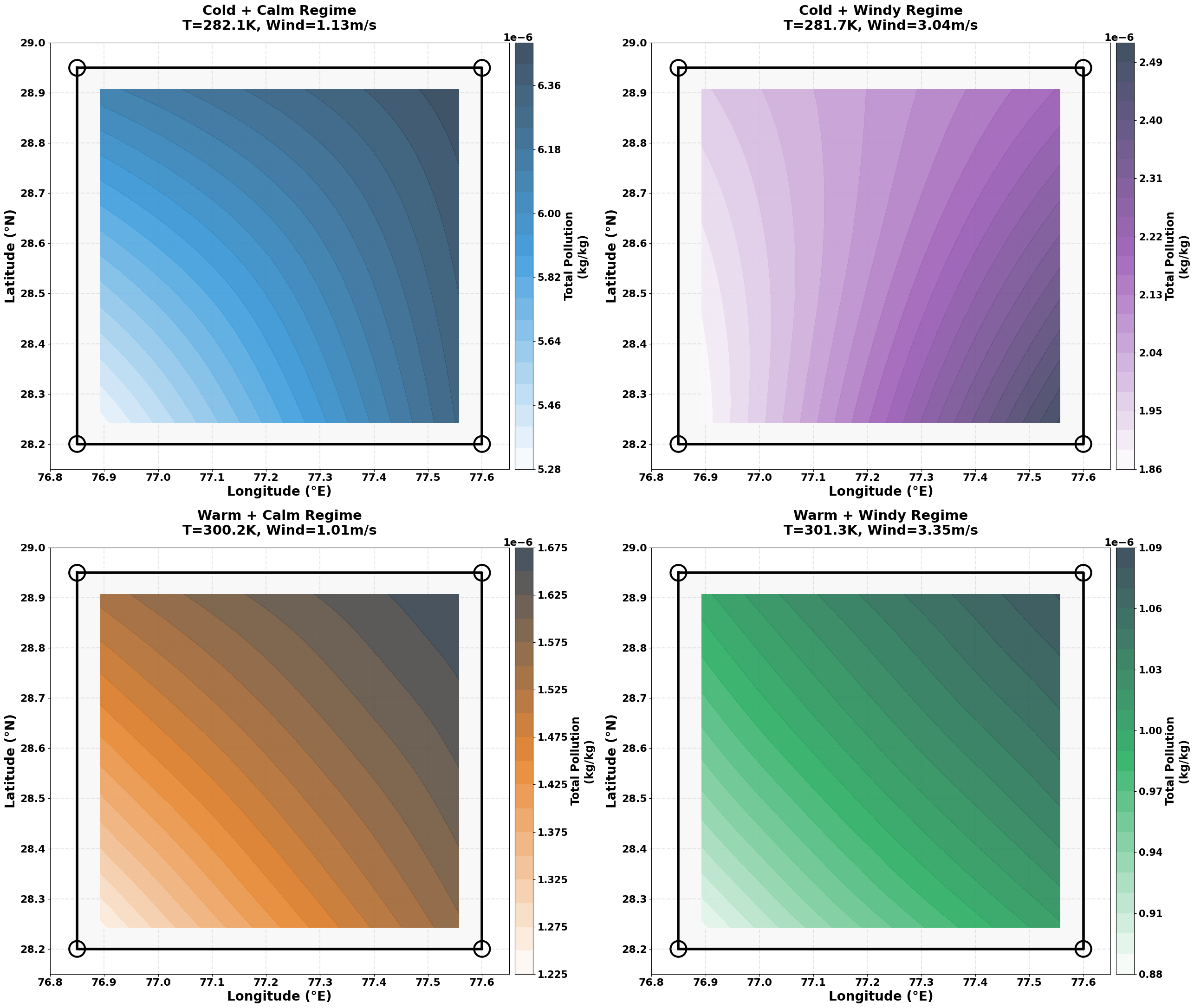}
\caption{Mean concentrations stratified by temperature and wind speed quartiles. Cold and calm conditions produce most severe pollution, exceeding six times warm and windy regime levels.}
\label{fig:weather_regimes}
\end{figure}

Combined meteorological regime analysis (Figure \ref{fig:weather_regimes}) isolates the interaction between temperature and wind speed. Cold and calm conditions ($T < 282$ K, wind $< 1.13$ m/s) produce the most severe pollution, with concentrations exceeding six times those under warm and windy regimes. Cold but windy conditions yield intermediate levels, indicating that horizontal dispersion partially compensates for the suppressed vertical mixing---a mechanistic distinction with practical implications for episode forecasting.

\begin{figure}[h]
\centering
\includegraphics[width=0.9\textwidth]{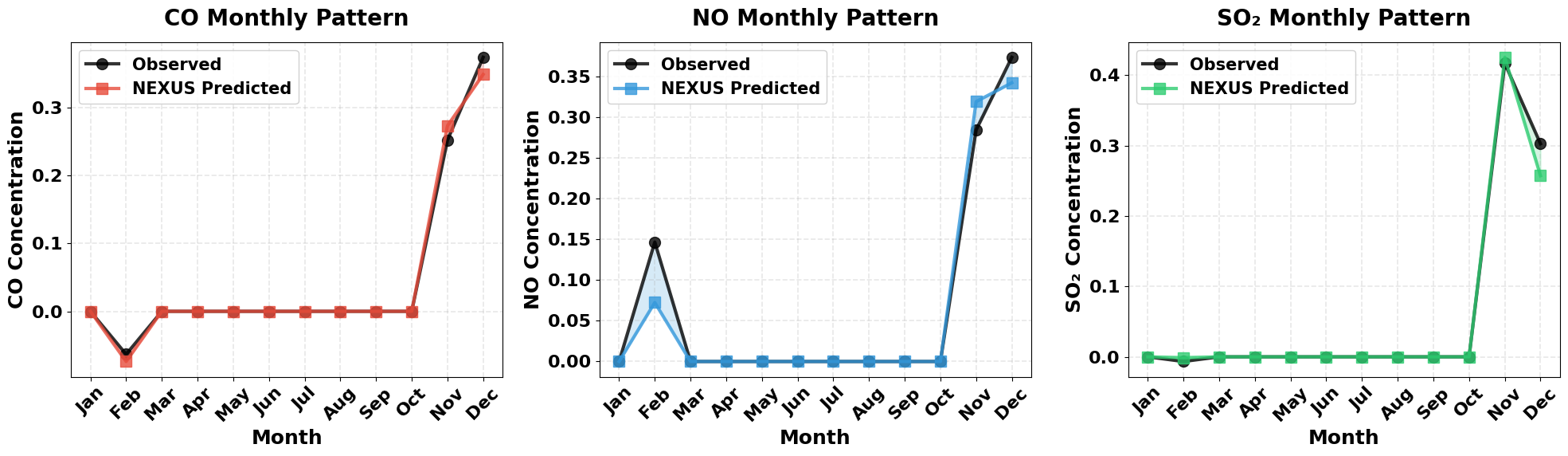}
\caption{Seasonal patterns in monthly means showing pronounced November-December peaks contrasted against March-September baseline values. \model\ accurately tracks these seasonal variations.}
\label{fig:monthly_patterns}
\end{figure}

Monthly patterns (Figure \ref{fig:monthly_patterns}) demonstrate the full amplitude of Delhi NCR's seasonal cycle. Concentrations remain low from March through September, with monthly means below 0.05 kg/kg, before the October onset marks the transition into the high-pollution season. November and December see the most severe conditions, with monthly means exceeding 0.35 kg/kg. \model\ predictions track these seasonal transitions with high fidelity, including the sharp onset in October that simpler models tend to smooth over.

\begin{figure}[h]
\centering
\includegraphics[width=0.9\textwidth]{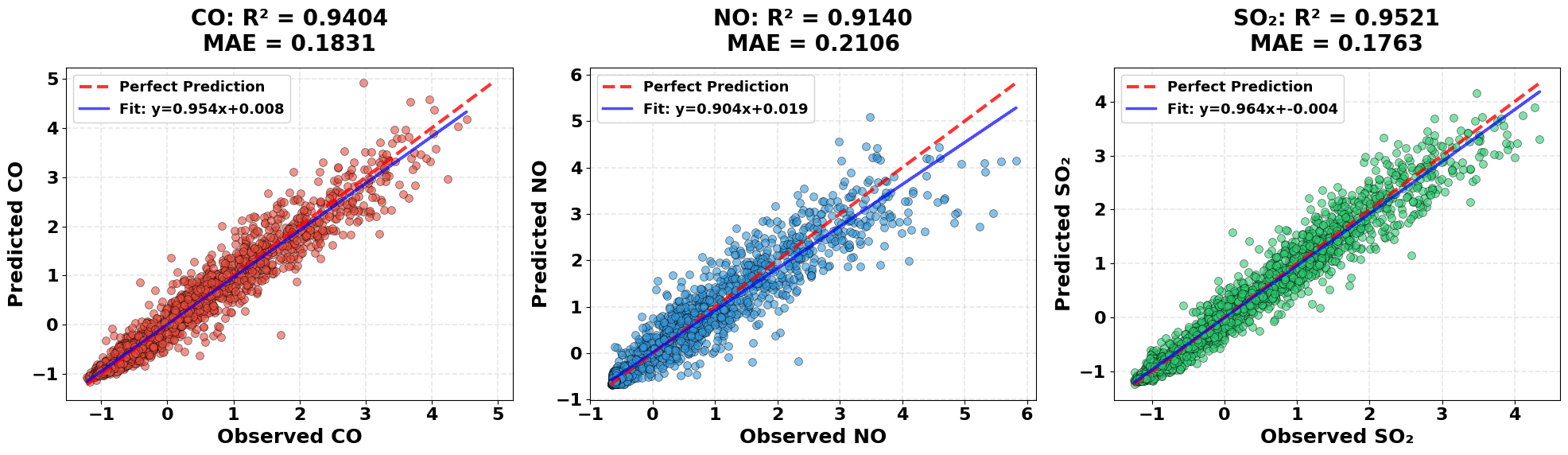}
\caption{Scatter plots comparing predicted and observed concentrations across test samples, showing tight clustering around perfect prediction line with minimal systematic bias.}
\label{fig:scatter_plots}
\end{figure}

Scatter plots (Figure \ref{fig:scatter_plots}) show tight clustering around the one-to-one line with minimal bias. Linear regression slopes are near unity for all three species---0.954 (CO), 0.904 (NO), 0.964 (SO$_2$)---and R$^2$ exceeds 0.91 across the board, confirming that the model neither systematically over- nor under-predicts at any concentration level.

\section{Ablation Study}

To understand which architectural choices drive performance, we conducted systematic ablation experiments removing one component at a time. Table \ref{tab:ablation} reports results alongside the full \model\ configuration.

Patch embedding contributes the largest single gain: removing it reduces average R$^2$ by 8.77\% while simultaneously increasing parameter count (from 18,748 to 24,320) and training time (from 25 to 38 minutes)---a result that underscores how patch-based compression improves both accuracy and efficiency together rather than trading one against the other. The low-rank projection accounts for a 5.72\% R$^2$ improvement, consistent with its role in filtering noise while preserving the dominant spatiotemporal modes most predictive of future concentrations; rank $r = 32$, selected from a grid over $\{16, 32, 64, 128\}$, struck the best balance between expressiveness and regularization. Removing the parallel NanoBlock pathways causes a 6.28\% R$^2$ drop, confirming that simultaneous multi-scale extraction captures information that any single pathway misses. Weighted spatial pooling contributes 4.09\%, reflecting the value of dynamically emphasizing monitoring sites most informative under prevailing wind conditions.

\begin{table}[h]
\centering
\caption{Ablation study showing contribution of each architectural component.}
\label{tab:ablation}
\renewcommand{\arraystretch}{1.3}
\begin{tabular}{@{}lcccc@{}}
\toprule
\rowcolor{gray!10}
\textbf{Configuration} & \textbf{Avg R$^2$} & \textbf{$\Delta$ R$^2$} & \textbf{Parameters} & \textbf{Training Time} \\
\midrule
\rowcolor{yellow!20}
\textbf{\model\ } & \textbf{0.9355} & \textbf{baseline} & \textbf{18,748} & \textbf{25 min} \\
w/o Patch Embedding & 0.8535 & -8.77\% & 24,320 & 38 min \\
w/o Low-rank Projection & 0.8820 & -5.72\% & 22,144 & 31 min \\
w/o NanoBlock Pathways & 0.8768 & -6.28\% & 14,592 & 19 min \\
w/o Weighted Pooling & 0.8972 & -4.09\% & 18,320 & 24 min \\
Single NanoBlock & 0.8654 & -7.49\% & 12,416 & 18 min \\
\bottomrule
\end{tabular}
\end{table}

Hyperparameter selection throughout followed systematic grid search on validation data. Patch size $p = 4$ and stride $s = 2$ (tested over $p \in \{2, 4, 8\}$, $s \in \{1, 2, 4\}$) proved optimal---larger patches lost the temporal resolution needed to resolve diurnal cycles, while smaller patches offered negligible computational savings. Hidden dimension $d' = 128$ (tested: $\{64, 128, 256\}$) provided sufficient capacity without excess parameters. Two NanoBlocks (tested: $\{1, 2, 3\}$) achieved the best accuracy-efficiency tradeoff. Dropout rate 0.1 and weight decay $10^{-4}$ gave the strongest generalization, and learning rate decay was found essential for fine-tuning as training progressed.

The parallel pathway design also outperforms sequential alternatives by 4.2\% in R$^2$. When pathways operate in parallel, CompactKernel, MicroConv, and FusionGate extract their respective feature types simultaneously; forcing a sequential order introduces an implicit hierarchy that has no physical motivation. Replacing NanoBlocks with LSTM layers (128 hidden units) reduced R$^2$ by 5.8\% while more than doubling parameters to 45,312. Substituting self-attention reduced R$^2$ by 3.4\% and expanded parameters to 156,928, confirming that for air quality forecasting---where most predictive signal is concentrated in the recent past and local spatial neighborhood---the marginal benefit of long-range attention does not justify its substantial computational cost.

\section{Conclusion}

This work introduces \model\, a compact neural architecture that simultaneously advances forecasting accuracy and computational efficiency for air quality prediction. Through three complementary design innovations---patch embedding, low-rank projections, and adaptive fusion---\model\ reduces parameters by 94\% relative to FEDformer while improving average R$^2$ by 6.95\%. The resulting model achieves R$^2$ exceeding 0.94 for CO, 0.91 for NO, and 0.95 for SO$_2$ with only 18,748 parameters and inference speeds six times faster than competing approaches.

Beyond raw predictive performance, the Delhi NCR analyses reveal pollution dynamics with direct policy relevance. The November--December concentration peaks---driven by the convergence of winter meteorological conditions and agricultural burning---point clearly to the need for anticipatory emergency protocols timed to the autumn transition rather than reactive responses to observed exceedances. The pronounced diurnal cycle, with morning and evening peaks far exceeding daily-average levels, underscores that policies calibrated to mean concentrations systematically underestimate peak exposures for vulnerable populations. The spatial gradients exceeding a factor of two during episodes make the case that neighborhood-level predictions, rather than regional averages, are necessary for equitable exposure assessment and intervention targeting.

The quantified meteorological relationships offer operational guidance of their own. Temperature's strong negative correlation with all three pollutants (r exceeding 0.5) establishes atmospheric stability---particularly overnight temperature minima that determine inversion intensity---as the dominant control on pollution severity. This finding implies that forecast quality depends critically on accurate temperature prediction, a tractable objective given the maturity of modern numerical weather prediction. Wind speed's role, while real, is secondary to vertical mixing in this terrain.

The efficiency gains achieved by \model\ prove as consequential as the accuracy improvements for real-world impact. Monitoring infrastructure in settings like Delhi NCR frequently runs on commodity hardware with limited GPU memory; models with sub-millisecond inference and 25-minute training cycles can be updated frequently and deployed broadly in a way that 90-minute, 298,000-parameter alternatives cannot. The interpretability analyses complement this operational advantage by translating model outputs into the mechanistic language of atmospheric science---identifying threshold conditions for episode onset, quantifying source region contributions, and characterizing the regime-specific behaviors that emergency response plans need to anticipate.

With 99\% of the global population living in areas exceeding WHO air quality limits, the gap between forecasting capability and operational deployment has measurable consequences for public health. \model\ demonstrates that purposeful architectural design---one that takes atmospheric data structure seriously rather than applying general-purpose sequence models off the shelf---can close this gap. Accuracy and efficiency need not be traded against each other; when an architecture is matched to the structure of the problem, both improve together.

\vspace{1cm}
\textbf{Acknowledgement.} We acknowledge the European Centre for Medium-Range Weather Forecasts (ECMWF) for providing ERA5 reanalysis data and the Copernicus Atmosphere Monitoring Service (CAMS) for pollutant concentration observations used in this study.

\bibliographystyle{abbrv}

\subsection*{Code Availability}
\noindent The complete experimental workflow is available as a GitHub repository:\\\url{https://github.com/msdsm03rampunitk-7606/NEXUS-Air-quality-forecasting-DELHI-NCR.git}
\end{document}